\documentclass[twocolumn]{rps-esrl2020}

\def\papername{\jobname}

\usepackage{amsmath,amssymb,amsfonts}
\usepackage{algorithmic}
\usepackage{graphicx}
\usepackage{textcomp}
\usepackage{xcolor}
\usepackage{array}

\usepackage[caption=false]{subfig}
\usepackage{makecell}
\usepackage{multirow}
\usepackage{tabularx}
\usepackage{float}

\setcitestyle{square}
\setcitestyle{numbers}

\begin{document}

\markboth{M. Beyer et al.}{Fault Injectors for TensorFlow: Evaluation of the Impact of Random Hardware Faults on Deep CNNs}

\twocolumn[

\title{Fault Injectors for TensorFlow: Evaluation of the Impact of Random Hardware Faults on Deep CNNs}
\author{%
Michael Beyer\textsuperscript{1}, 
Andrey Morozov\textsuperscript{2}, 
Emil Valiev\textsuperscript{1}, 
Christoph Schorn\textsuperscript{3}, 
Lydia Gauerhof\textsuperscript{3}, 
Kai Ding\textsuperscript{4}, 
Klaus Janschek\textsuperscript{1}%
}
\address{%
\textsuperscript{1}Institute of Automation, Technische Universität Dresden, Germany.\\ 
\textsuperscript{2}Institute of Industrial Automation and Software Engineering, University of Stuttgart, Germany.\\
\textsuperscript{3}Robert Bosch GmbH, Corporate Research, Renningen, Germany.\\
\textsuperscript{4}Bosch (China) Investment Ltd., Corporate Research, Shanghai, China.\\
\email{
\{michael.beyer3, emil.valiev\}@mailbox.tu-dresden.de, andrey.morozov@ias.uni-stuttgart.de, \{christoph.schorn, lydia.gauerhof\}@de.bosch.com, kai.ding@cn.bosch.com, klaus.janschek@tu-dresden.de
}}

\begin{abstract} 
Today, Deep Learning (DL) enhances almost every industrial sector,
including safety-critical areas. The next generation of safety standards will
define appropriate verification techniques for DL-based applications and propose
adequate fault tolerance mechanisms.
DL-based applications, like any other software, are susceptible to common random
hardware faults such as bit flips, which occur in RAM and CPU registers. Such
faults can lead to silent data corruption. Therefore, it is crucial to develop
methods and tools that help to evaluate how DL components operate under the
presence of such faults. 
In this paper, we introduce two new Fault Injection (FI) frameworks InjectTF and
InjectTF2 for TensorFlow 1 and TensorFlow 2, respectively. Both frameworks are
available on GitHub and allow the configurable injection of random faults into
Neural Networks (NN). In order to demonstrate the feasibility of the frameworks,
we also present the results of FI experiments conducted on four VGG-based
Convolutional NNs using two image sets. The results demonstrate how random bit
flips in the output of particular mathematical operations and layers of NNs
affect the classification accuracy. These results help to identify the most
critical operations and layers, compare the reliability characteristics of
functionally similar NNs, and introduce selective fault tolerance mechanisms.
\end{abstract}

\keywords{Fault Injection, Random Hardware Faults, Deep Learning, Fault Tolerance, Reliability}

]


\section{Introduction}
\label{sec:intro}

Nowadays, Artificial Intelligence (AI) is exploited all over the industry, including safety-critical applications in the automotive, aerospace, and medical sectors.
As part of AI, computer vision is an essential task in autonomous driving, robotics, and healthcare.
The autonomous driving use cases vary from the recognition of pedestrians and traffic signs to the identification of lane markings.
In robotics, AI helps with object classification and visual navigation.
Furthermore, AI supports doctors in patient monitoring, diagnosing tasks, and development of treatment protocols for personalized medicine.
The mentioned applications are based on image recognition using
Convolutional Neural Networks (CNN).
CNNs are considered to be state of the art for this task as this type of network takes the lead in the ImageNet Large Scale Visual Recognition Challenge (ILSVRC)~\cite{ILSVRC15}.
One of the commonly used open-source machine learning frameworks for the development of neural networks is \mbox{TensorFlow~\cite{tensorflow2015-whitepaper2}}, which has proven to be a stable and reliable platform.
The flexible architecture of TensorFlow allows for easy deployment across a variety of platforms, including embedded devices.
TensorFlow version 1.0 has been released by the Google Brain team in Februray
2017, the second version 2.0 became officially available in September 2019.
Currently, both versions are in use.

Neural networks, like any software, are susceptible to common random hardware faults such as bit flips.
Lowering the voltage, increasing heat, radiation, electromagnetic induction, or other negative environmental impacts can corrupt CPU or RAM and change the value of a stored variable.
This affects the classification accuracy of the network.
Safety-critical applications that utilize AI-based methods have to prove their reliability according to industrial safety standards.
Therefore, investigating the effects of such soft errors is crucial to ensure the correct operation of the AI application under the effect of such errors.

\textbf{Contribution:}
This paper presents two FI frameworks for TensorFlow 1 and TensorFlow 2.
The frameworks allow the user to model soft errors during the execution of a neural network, analyze the effects of such errors, and identify the most critical parts of the network.

To demonstrate the feasibility of the developed frameworks, we conducted four large-scale FI experiments.
These experiments yield the classification accuracy of four neural networks under varying circumstances.
The tested networks include two VGGs (VGG16 and VGG19)~\cite{simonyan2014deep}, an
open-source VGG-based network, and a simple CNN based on general design guidelines.
For the experiments the German Traffic Sign Recognition Benchmark (GTSRB)~\cite{journals/nn/StallkampSSI12} and ImageNet~\cite{ILSVRC15} datasets have been used.
The results show which erroneous operations or layers have the largest impact on the accuracy, allowing the introduction of efficient selective fault protection mechanisms.

The remainder of the paper is structured as follows.
First, the developed FI frameworks for TensorFlow $1$ and
$2$ are presented in Section~\ref{sec:fifw}.
Section~\ref{sec:experiments} gives an overview of the experimental setup.
The results are shown and discussed in Section~\ref{sec:results}.
Afterward, an overview of related work is given in Section~\ref{sec:relatedWork}.
Finally, this work is concluded in Section~\ref{sec:conclusion}.


\begin{figure*}[htbp]%
\centering
  \subfloat[Working principle of the fault injection framework
	  InjectTF\@. The operation selected for injection is shown in orange. Source code
  available at https://github.com/mbsa-tud/InjectTF\,.
	\label{fig:fifw:injectTF:workingPrinciple}
]{\includegraphics[width=0.923\linewidth]{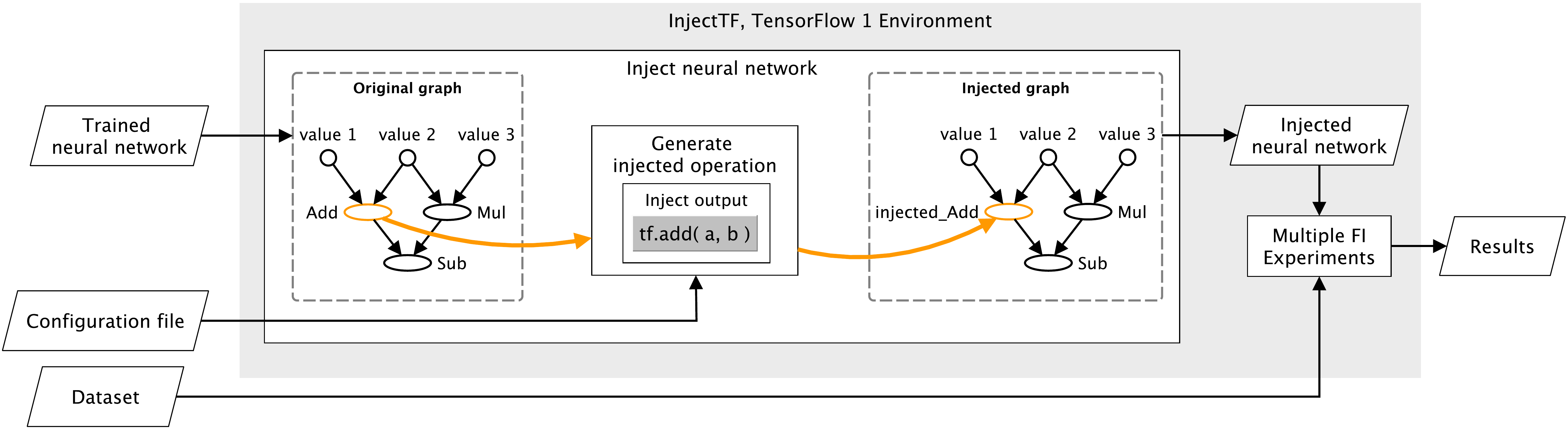}}\\%
\subfloat[Working principle of the fault injection framework InjectTF2\@. The layer
	  selected for injection is shown in orange. Source code available at
	  https://github.com/mbsa-tud/InjectTF2\,.
	\label{fig:fifw:injectTF2:workingPrinciple}]{%
	\includegraphics[width=0.923\linewidth]{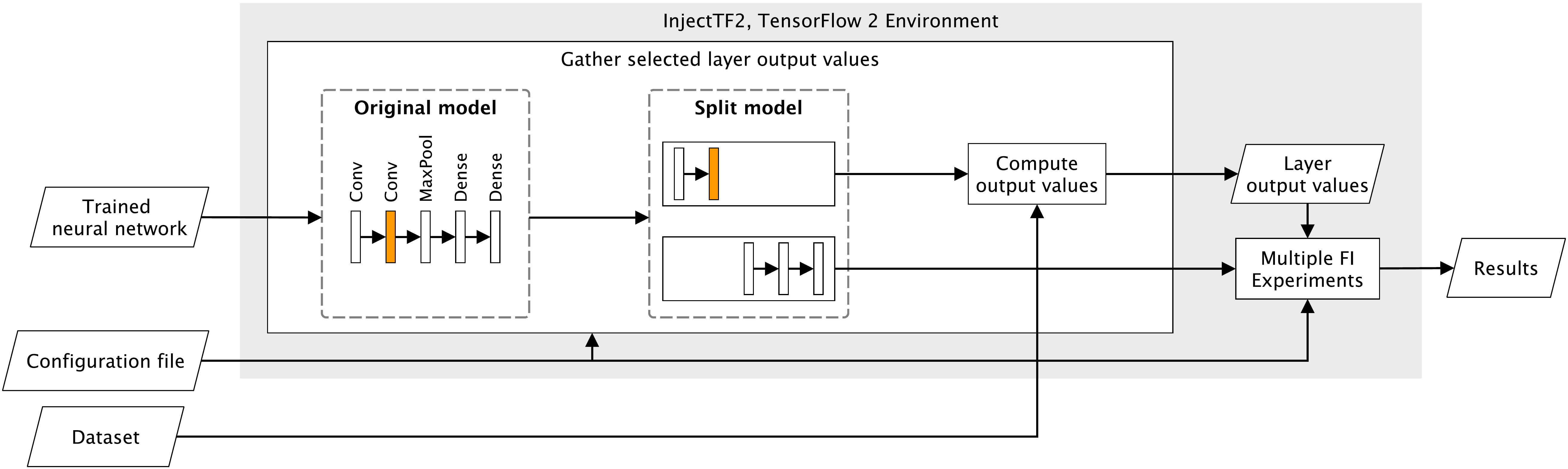}}%
	\caption{Working principle overview of both fault injection frameworks.}
	\label{fig:fifw::workingPrinciple}
\end{figure*}

\section{Fault injection frameworks}
\label{sec:fifw}

We present two new FI frameworks.
\emph{InjectTF} for TensorFlow 1 and \emph{InjectTF2} for TensorFlow 2.
The key capabilities of both frameworks are outlined in Table~\ref{tab:fifwComp}.
They are designed in a modular way and can therefore easily be expanded with additional features.
A high-level overview of the first framework, InjectTF, was already presented in our two-page abstract~\cite{8990333}.

\begin{table}%
\begin{center}
\tbl{Comparison of the two fault injection frameworks}{%
\begin{tabular}{|l|c|c|}
\hline
& 	\textbf{InjectTF}	& \textbf{InjectTF2}\\
\hline
\makecell[c]{TensorFlow\\version} & \makecell[c]{Version $1$,\\low-level\\TF API} & \makecell[c]{Version $2$,\\high-level\\Keras API}\\
\hline
\makecell[c]{FI style} & Operation-wise & Layer-wise\\
\hline
\makecell[c]{Supported\\fault types} & \makecell[c]{Zero, random\\value, random\\bit flip} & \makecell[c]{Specific or\\random\\ bit flip}\\
\hline
\makecell[c]{Supported\\operations/\\model structures} & \makecell[c]{Add, Sub,\\Mul, ReLU$^{\mathrm{1}}$} & Sequential$^{\mathrm{2}}$ \\
\hline
\multicolumn{3}{l}{$^{\mathrm{1}}$The operation list can be easily extended by the user.}\\
\multicolumn{3}{l}{$^{\mathrm{2}}$Parallel structures will be supported in the next version.}
\end{tabular}}
\label{tab:fifwComp}
\end{center}
\end{table}

In this full paper we give a detailed and up-to-date description.
InjectTF injects errors into the outputs of individual operations of neural networks
and has been developed using the low-level TensorFlow 1 API\@.
TensorFlow 2.0 brought significant changes to its Python API\@.
With the integration of the high-level Keras API, designing neural networks becomes more abstract
as the paradigm has been shifted from the operation-wise development to a layer-wise.
Now the layers are the basic building blocks instead of the individual operations that compose a layer.
InjectTF could not be adopted to TensorFlow 2.0, therefore a complete
redesign was required.
We follow this trend of increasing abstraction and focus on injecting faults into the output of layers instead of individual operations.
Since the updated version of InjectTF fundamentally changes the behavior of the framework, it is developed under a new name: \emph{InjectTF2}.
Both frameworks, including examples, are available on GitHub\footnote{Source code available at https://github.com/mbsa-tud}.

\subsection{InjectTF}
\label{sec:fifw:injectTF}

\noindent
\textbf{Architecture:}
InjectTF is built on top of the standard TensorFlow $1$ library and takes a configuration file as well as a trained neural network as inputs to create a fault injected counterpart of the original TensorFlow model.
The network has to be provided either as a checkpoint and a meta file or as a frozen graph.

\noindent
\textbf{Configuration:}
In the configuration file, the user lists TensorFlow operation type(s) along with FI probabilities.
It is possible to choose between three types of faults that will be injected with the chosen probability into the output of the listed operation(s): (i) change one random value of the operation's result to zero, (ii) change one random value of the operation's result to a random number, or (iii) flip one random bit of a random value of the operation's result.

\noindent
\textbf{Working principle:}
The process of creating the injected model is sketched in the middle of Figure~\ref{fig:fifw:injectTF:workingPrinciple}.
Here the TensorFlow operation \emph{Add} is selected for injection.
InjectTF iterates over all operations within the graph of the ``original'' network.
All operations that are not listed by the user are just copied to a new ``injected'' graph.
However, if an operation type is in the list, then an injected counterpart of that operation is generated and added to the injected graph.
The new operation is essentially the wrapped ``original'' TensorFlow operation.
Therefore, when executing the injected network, only the result of the
underlying computation is altered according to the parameters defined in the configuration file.
During runtime, both neural networks can be used for inference.

\subsection{InjectTF2}
\label{sec:fifw:injectTF2}

\noindent
\textbf{Architecture:}
InjectTF2 is built on top of the standard TensorFlow $2$ library and requires a configuration file, a trained neural network, and a dataset as input.
In Figure~\ref{fig:fifw:injectTF2:workingPrinciple} the inputs required for
initialization of the framework are shown on the left side.
The network and the dataset should be provided as a HDF5 model and a TensorFlow dataset, respectively.
The latter is used to save the intermediate output values of the layer that is to be injected with errors in the experiments.
As of now, the framework only works with sequential models that have been developed with the high-level Keras API\@.

\noindent
\textbf{Configuration:}
In the configuration file, the layer that is to be injected is specified along
with the injection probability.
Currently, only bit flips can be injected by the framework.
However, it is possible to choose between flipping a random bit or a specific bit of a random value of a layer's output.

\noindent
\textbf{Working principle:}
In the middle of Figure~\ref{fig:fifw:injectTF2:workingPrinciple}, the working principle of InjectTF2 is shown.
When initializing the framework, we execute the model up to the layer where the errors are to be injected, using the provided data set.
The output values of the injected layer are collected and stored.
Afterward, we continue the analysis from the selected layer onward using the gathered values.
Errors are injected into the stored values of the selected layer according to the parameters defined in the configuration file.

Splitting the model and executing the two resulting parts separately
drastically reduces the execution time of the experiments, since the network is not executed from bottom to top each time.
During runtime, however, only the original model exists.
Compared to InjectTF, InjectTF2 is therefore more dynamic, since it neither alters the structure of the network nor creates a new model with injected faults.
However, a substantial amount of memory is required to store the intermediate values of the selected layer.
Depending on the layer and dataset size, this can quickly reach $100\,\text{GB}$.

\section{Experimental setup}
\label{sec:experiments}

Three image classification experiments with CNNs have been conducted using the two presented FI frameworks.
An overview of the experimental setup is given in Table~\ref{tab:expSetup}.
The first experiment has been done with InjectTF and an open-source traffic sign classifier~\cite{vamsi} that is based on the VGGNet\@. For the second one, we developed a simple CNN-based traffic sign classifier with TensorFlow $2$ and injected faults into it with InjectTF2.
In the last experiment, we concentrate on publicly available networks integrated into TensorFlow, in order to identify common vulnerabilities of neural networks.
Since previous experiments were based on the VGGNet, we use pre-trained VGG16 and VGG19 networks and inject errors into those with InjectTF2.

To ensure the statistical confidence of our experiments, we compute the Cumulative Moving Average (CMA) for the resulting classification accuracy in each experiment.

\begin{table*}[htbp]
\tbl{Overview of the experimental setup.}{%
\begin{tabular}{|l|c|c|c|}%
\hline
  & 	\textbf{Operation-wise FI}	&
                                                  \multicolumn{2}{c|}{\textbf{Layer-wise
                                                  FI}} \\
\hline
\makecell[l]{FI framework}    & InjectTF     & InjectTF2 & InjectTF2\\
\hline
\makecell[l]{Neural network} & VGG-based~\cite{vamsi} & Custom CNN & VGG16, VGG19\\
\hline
Dataset & GTSRB & GTSRB & ImageNet \\
\hline
Experiment & \makecell{Evaluate classification\\accuracy with a
  varying\\probability for FI 
			 } & \makecell{Evaluate classification accuracy\\with a varying probability
						  for FI\\and with
						  $100\,\%$ probability for FI} &
																 \makecell{Evaluate
																 classification
																 accuracy\\with
																 $100\,\%$
																 probabilty for
																 FI} \\
\hline
\makecell[l]{Injected\\operations/layers} & \makecell[c]{Operations Add, Sub,\\and Mul (separately)}& All layers
															 (separately) & All
																			layers
  (separately)\\
\hline
\end{tabular}}
\label{tab:expSetup}
\end{table*}

\subsection{Operation-wise FI Experiment}
\label{sec:experiments:issre}

\noindent
\textbf{Neural network:}
In this experiment, a VGG-based traffic sign classifier is used.
It consists of a localization network with a spatial transformer and a VGG-like network for classification.
The network has been trained on an augmented GTSRB dataset that was split into three subsets for training, testing, and validation.
The subsets contain $129\,100$, $12\,630$, and $4\,410$ images with $32\times32$ RGB pixels that belong to $43$ different types of road signs.
After training the network can classify traffic signs with an accuracy
of approximately $96\,\%$.

The network's most common TensorFlow operations are \emph{Add}, \emph{Sub}, and \emph{Mul}, which together form part of the main components of the Parametric ReLU (PReLU) activation function, that is used in most layers of the network, see Figure~\ref{fig:results:issre:PReLUTensorBoard}.

\noindent
\textbf{FI setup:}
Each of the previously mentioned operations is investigated in a set of separate experiments, whereby bit flips are injected with a varying probability into the output of the respective operation.
Depending on the probability of FI, one random bit of a random element of the operation's output is flipped.
Since the selected operations occur multiple times in the network, multiple bit flips might be injected during inference.
Each of the selected probabilities is tested $100$ times.
The classification accuracy is determined in each experiment using the GTSRB testing subset with the corresponding ground truth.
This helps to evaluate the effect of random faults in those operations on the performance of the network.

\subsection{Layer-wise FI Experiments}
\label{sec:experiments:layerWise}

\noindent
\textbf{Neural networks:}
In the layer-wise FI experiments, we use three different neural networks.
The first is a self-developed simple CNN based on general design guidelines.
It consists of $12$ layers and uses the ReLU activation function for all
convolutional and dense layers except the last, which uses the Softmax function.
As the network in the first experiment, our simple CNN has been trained on the augmented GTSRB dataset and is capable of classifying traffic signs with an accuracy of approximately $96\,\%$.
Additionally, we conduct experiments with pre-trained VGG16 and VGG19 networks with ImageNet weights.

\noindent
\textbf{FI setup:}
The experiments using our simple CNN network can be further divided into two categories: (i) Stochastic experiments with varying probability and (ii) deterministic experiments with a fixed $100\,\%$ probability of error injection.

In the first category, the performance of the network is evaluated by calculating the classification accuracy using the GTSRB testing subset's ground truth.
Similar to the experiment of Section~\ref{sec:experiments:issre}, each layer of the network is analyzed in a set of separated experiments, whereby one bit flip is injected with a varying probability into the output of the respective layer.
Depending on the probability of error injection, one random bit of one random element of the layer's output is flipped.
Each of the selected probabilities is tested $100$ times.

In the second category, each layer of the network is investigated in a set of
separate experiments, whereby the probability for injecting a random bit flip into the output of the respective layer is $100\,\%$.
The performance (accuracy) is determined according to the network's \emph{golden run} predictions, i.e., the predictions of the neural network without FI.

In the experiments with the VGG networks, the performance of both networks is determined using a random sample of $5000$ $224\times224$ RGB images from the $2012$ ImageNet testing subset.
As in the second category mentioned above, a set of $100$ bit flip injection experiments is conducted for each layer.
In these experiments, one random bit in one random element of the layer's output is flipped.
The ground truth of the ImageNet test subset is not available. Thus, the only option is to compare the performance of the networks to the predictions of the neural network without faults.
We call this prediction the \emph{golden run}.

\section{Results of the Experiments}
\label{sec:results}

\subsection{Operation-wise FI Experiment}
\label{sec:results:issre}

\begin{figure}[htpb]%
\centering
\includegraphics[width=0.886\linewidth]{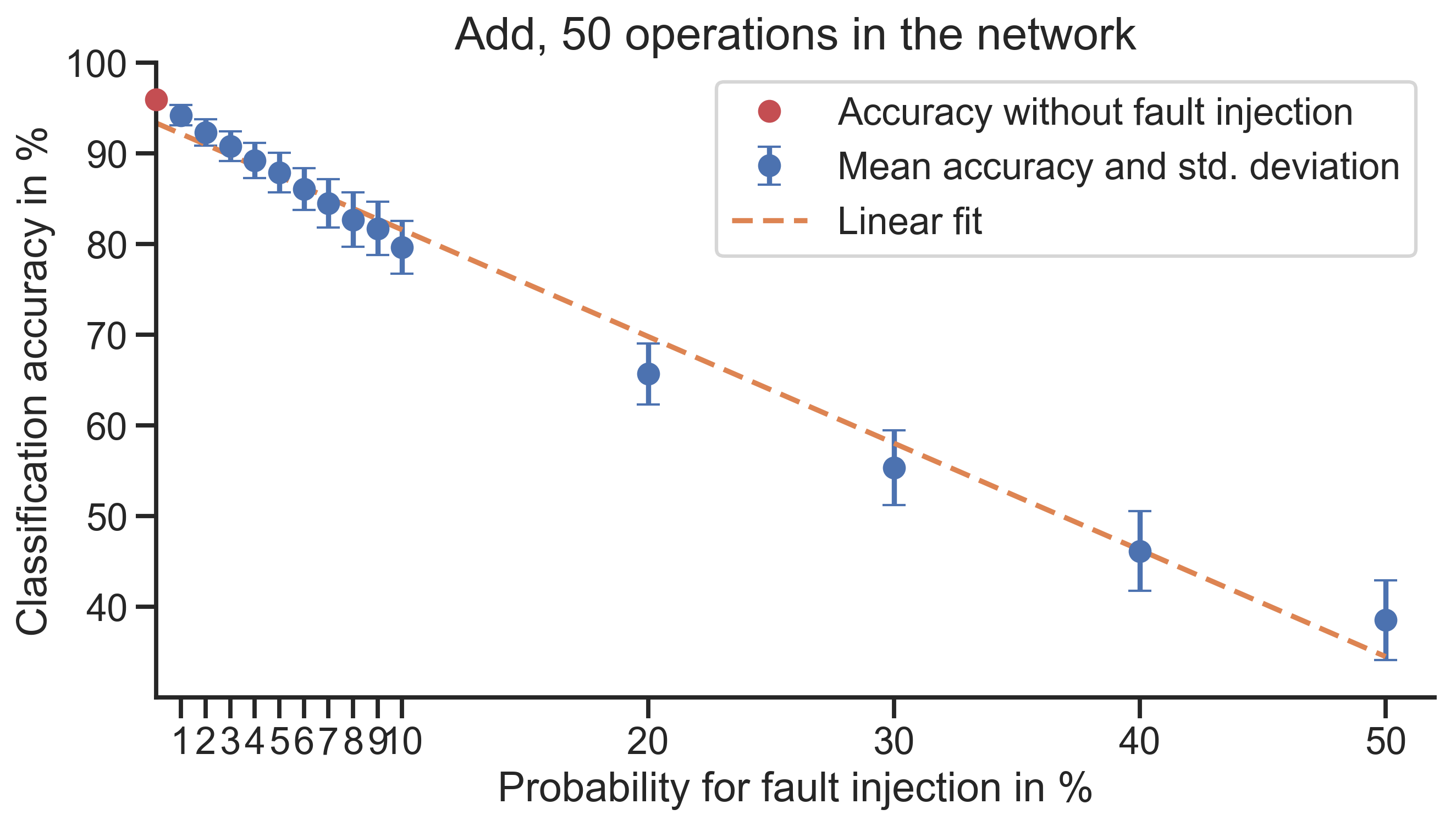}\\%
\includegraphics[width=0.886\linewidth]{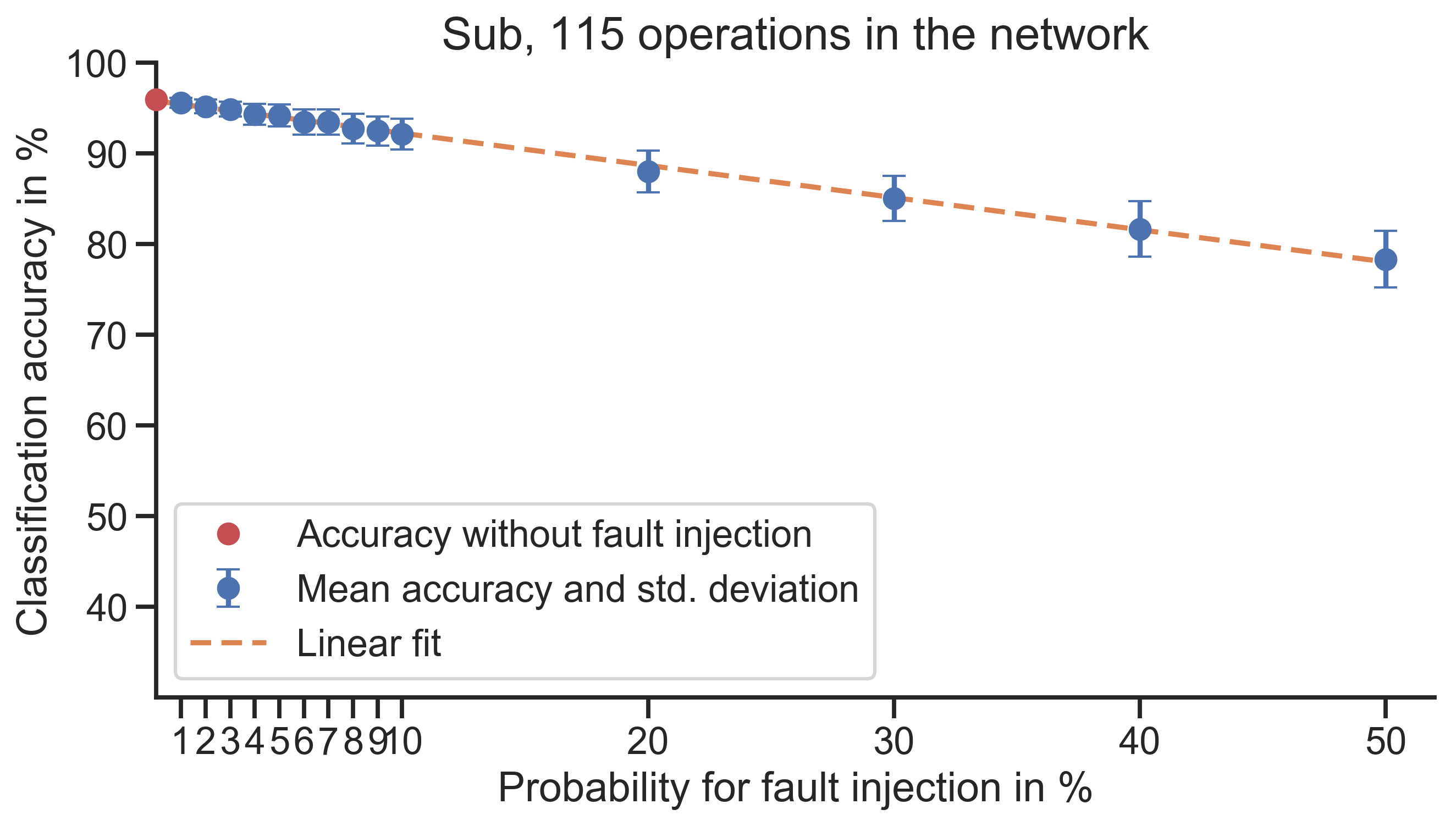}\\%
\includegraphics[width=0.886\linewidth]{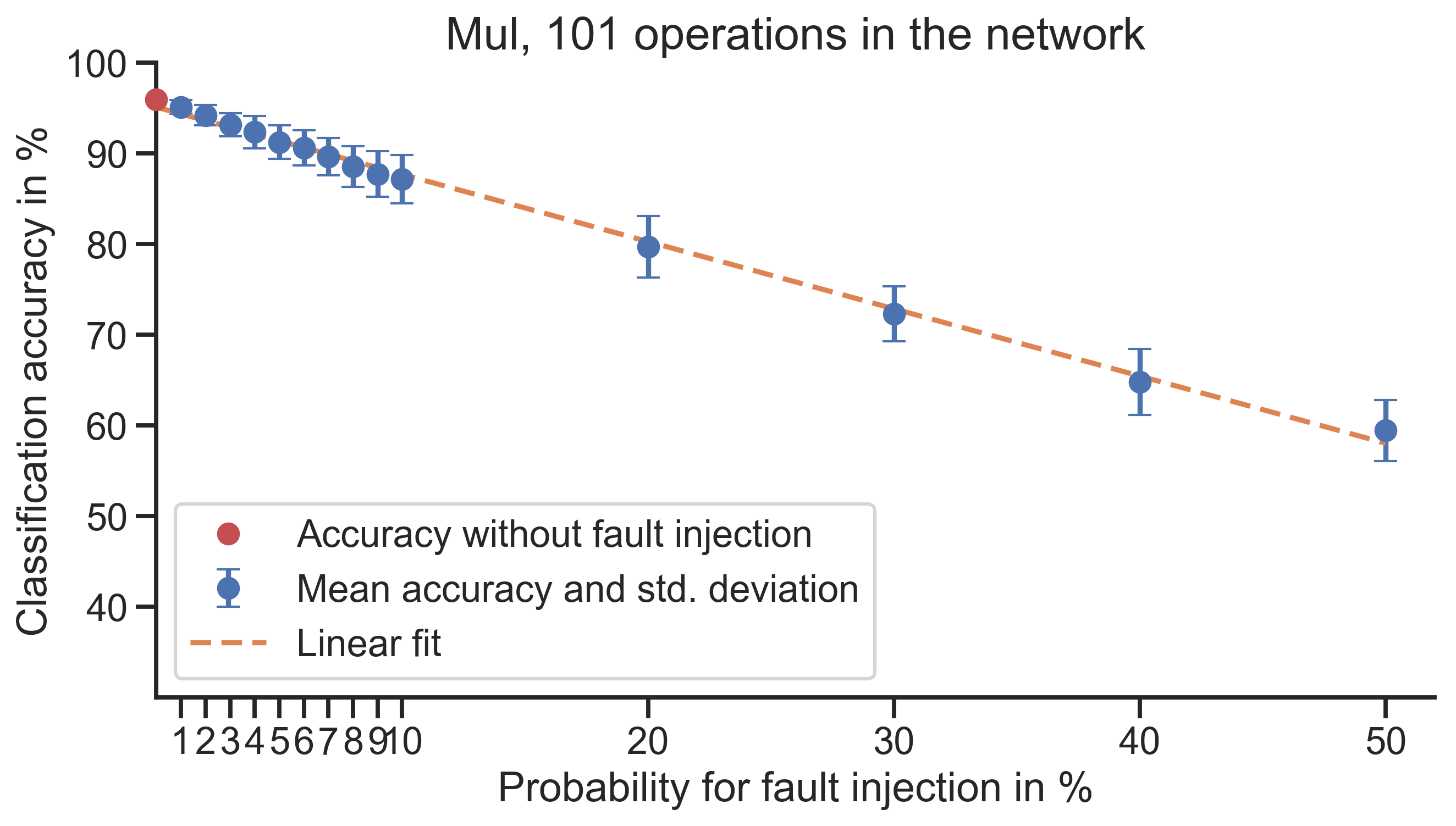}\\%
\caption{Results of the operation-wise bit flip FI experiments into the VGG-based CNN from Section~\ref{sec:experiments:issre}. It has been trained on an augmented GTSRB dataset~\cite{8990333} \textcopyright{} 2019 IEEE.}
\label{fig:results:issre:issreResults}
\end{figure}

Figure~\ref{fig:results:issre:issreResults} shows the results of the operation-wise FI experiments for each of the operations mentioned in Section~\ref{sec:experiments:issre}.
The varying FI probabilities are denoted on the X-axes.
The mean classification accuracy and standard deviation for the $100$ individual experiments are shown in blue.
The reference classification accuracy without FI is shown in red.

\begin{figure}[htpb]
  \centering
	\includegraphics[width=0.296\linewidth]{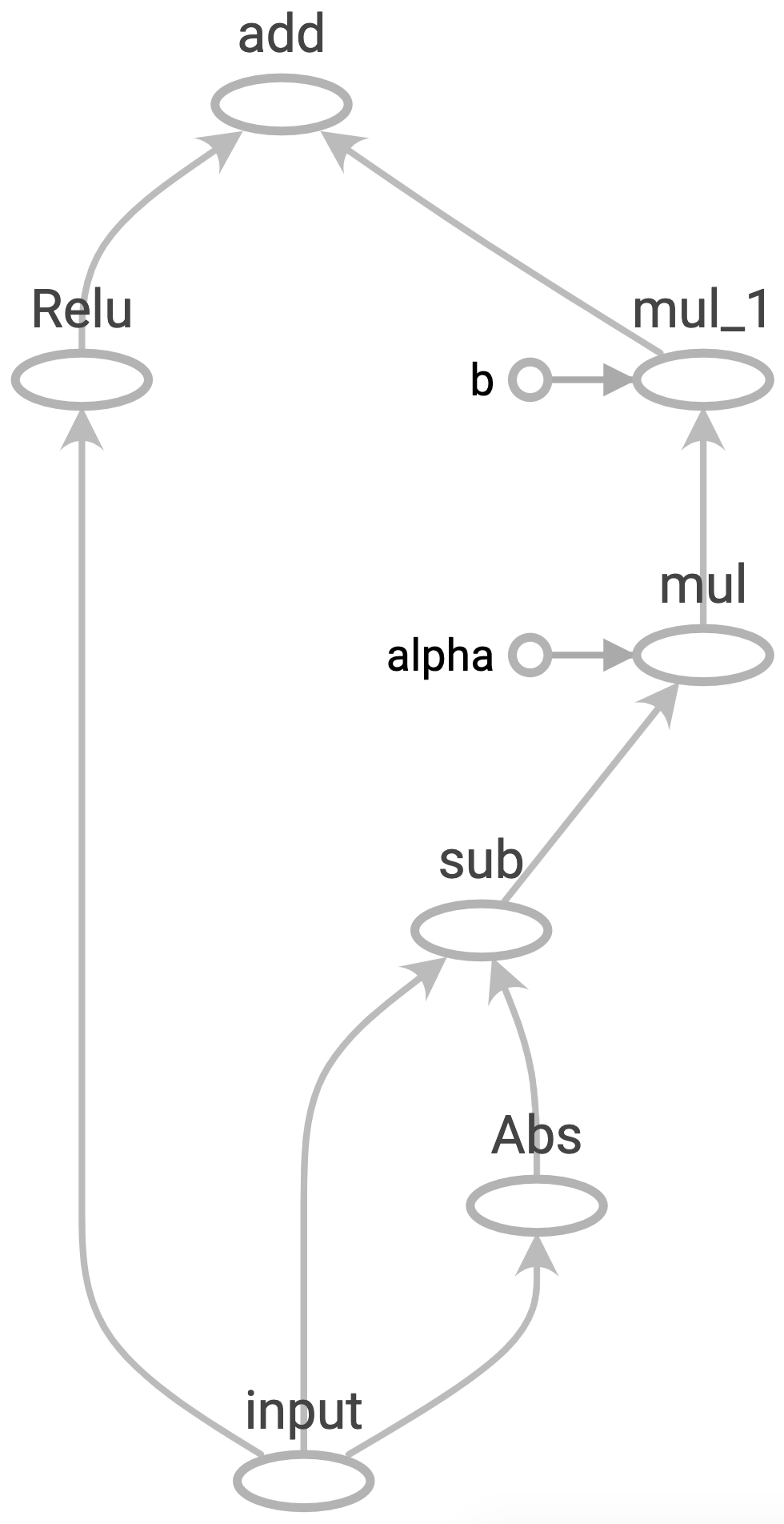}
	\caption{Visualization of the PReLU activation function implementation with TensorFlow
      operations in TensorBoard.}
	\label{fig:results:issre:PReLUTensorBoard}
\end{figure}

The resulting classification accuracy decreases linearly with an increasing probability for FI for all three investigated operations.
However, faults in some operations affect the resulting classification accuracy more than others.
The second plot in Figure~\ref{fig:results:issre:issreResults} shows that for this network, the classification accuracy decreases the least in case of errors in the \emph{Sub} operation.
Intuitively, one would expect a multiplication to have a significant impact.
As the third plot in Figure~\ref{fig:results:issre:issreResults} shows, faults in the \emph{Mul} operation indeed have a considerable influence on the resulting classification accuracy.
However, compared to the first plot in Figure~\ref{fig:results:issre:issreResults}, the \emph{Add} operation has a much more significant impact, although there are fewer \emph{Add} operations in the network.
This unexpected result can be explained by investigating the implementation of the PReLU activation function, which is used throughout the neural network used in this experiment.
Figure~\ref{fig:results:issre:PReLUTensorBoard} shows the implementation of the PReLU function with TensorFlow operations.
The graph is executed from bottom to top and can be split into two branches.
The left part consists of the standard TensorFlow ReLU operation, which corresponds to the positive domain of the PReLU activation function.
In the right part the negative domain of the function is realized by a combination of multiple operations.
\emph{alpha} is a trainable parameter, \emph{b} is a constant value equal to $0.5$.
The result of this branch has to be zero for all positive values.
Therefore, by subtracting the input's absolute value from the input and
multiplying the overall result with $0.5$, the desired result is obtained.
The sum of both branches determines the output of each layer with a PReLU activation function.
Consequently, errors in the \emph{Add} operation can have a significant impact on the classification accuracy. 
Also, these errors can cause important features to disappear when a max-pooling layer follows an erroneous \emph{Add} operation.
Therefore, this type of operation can be considered as the most critical one in this network.
The CMA for all tested operations converges to a finite value.
Thus, the sample size for the experiment is considered to be sufficient.

\vspace{6pt}
\noindent
\textbf{Key points of the operation-wise experiments:}%
\begin{itemlist}
\item The experiments help to identify the most critical operation type in the network under test.
\item \emph{Add} is the most critical investigated operation, presumably because of the PReLU function's structure.
\item Efficient selective operation-wise fault tolerance mechanisms can be introduced, based on the obtained information.
\item The classification accuracy tends to drop linearly with an increasing FI probability.
\end{itemlist}

\subsection{Layer-wise FI Experiments}
\label{sec:results:layerWise}

\begin{figure*}[htb]
  \centering
  \subfloat[	Resulting classification accuracies with a varying probability for fault injection.
	\label{fig:results:emil:simpleCnn:layerwiseVarying}]{%
	\includegraphics[width=0.46\linewidth]{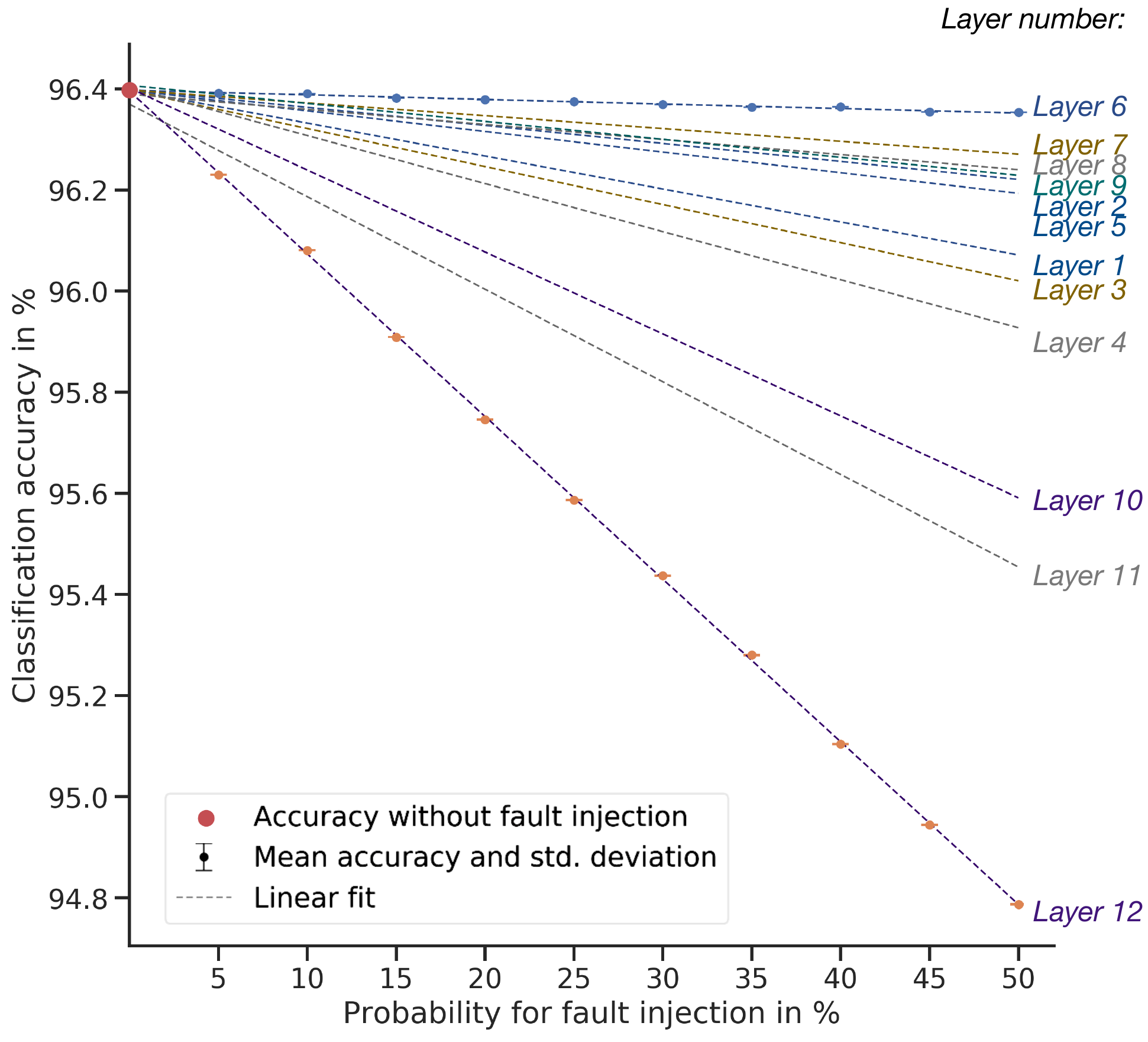}}
  \quad
  \subfloat[Resulting classification accuracies with a $100\,\%$ probability
      for fault injection.
	\label{fig:results:emil:simpleCnn:layerwise100}]{%
	\includegraphics[width=0.46\linewidth]{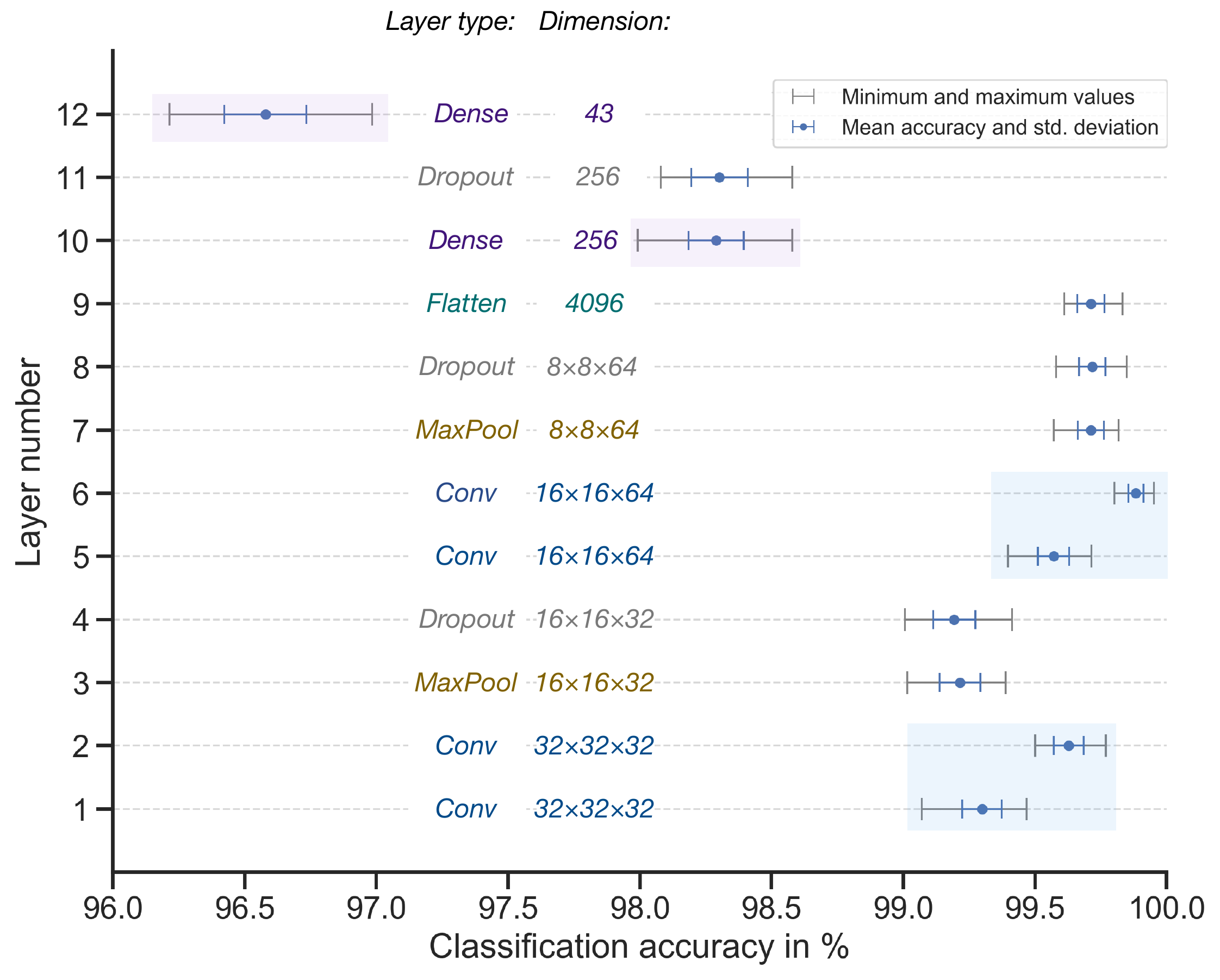}}
	\caption{Resulting classification accuracies for layer-wise bit flip
	  injection using InjectTF2. The network used is our self-developed CNN traffic sign classifier from
	  Section~\ref{sec:experiments:layerWise} which has been trained on an augmented GTSRB data set.}
	\label{fig:results:emil:simpleCnn}
\end{figure*}

Figure~\ref{fig:results:emil:simpleCnn:layerwiseVarying} shows the results of the first category of experiments on our simple CNN, described in Section~\ref{sec:experiments:layerWise}.
The FI probabilities are denoted on the X-axis.
The reference classification accuracy without FI is shown in red.
For each layer, the resulting mean classification accuracy and standard deviation for each of the individual $100$ experiments per probability have been computed.
Overall, the standard deviation is negligibly small across all tested
probabilities.
For the sake of readability, we only show the mean classification accuracy
and standard deviation for the ``best'' and ``worst'' layer (i.e. layer $6$ and layer $12$).
For all other layers we show only the linear fit, which is color-coded according to the layer type.
The corresponding layer numbers for each linear fit are shown on the right side.

Compared to the results of the operation-wise FI, discussed in Section~\ref{sec:results:issre}, the overall drop in classification
accuracy is significantly less.
This can be explained by the fact that in the operation-wise FI experiments, more than one fault can be injected during the execution of the network.
Consequently, the results presented in Figure~\ref{fig:results:emil:simpleCnn:layerwiseVarying} are not as severe as those in Section~\ref{sec:results:issre}.
Nevertheless, the linear relationship between the probability for FI and the resulting classification accuracy is still apparent.

\begin{figure*}[htb]
  \centering
  \subfloat[Resulting classification accuracies for the VGG16 CNN.
	\label{fig:results:vgg:vgg16}]{%
	\includegraphics[width=0.46\linewidth]{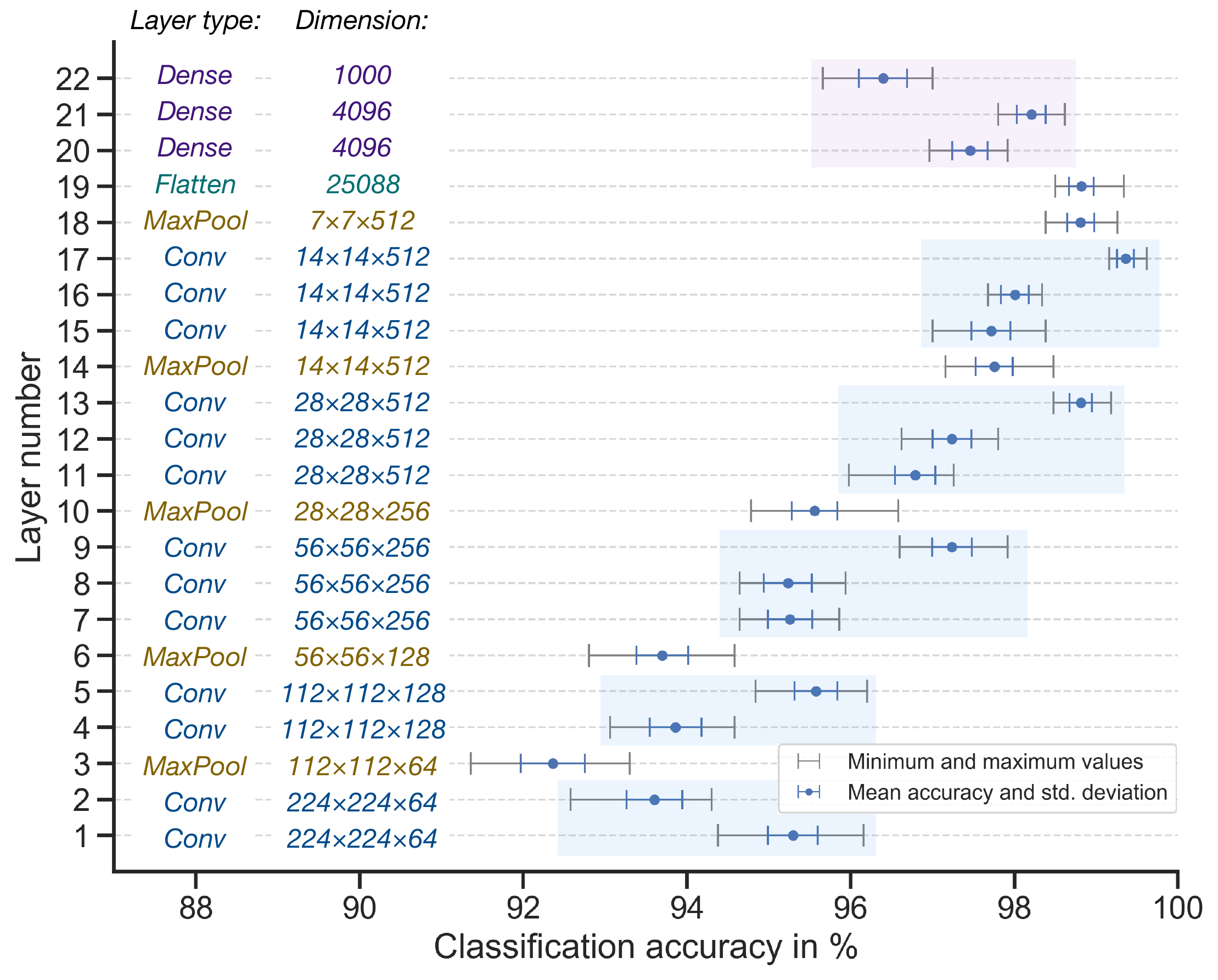}}
  \quad
  \subfloat[Resulting classification accuracies for the VGG19 CNN.
	\label{fig:results:vgg:vgg19}]{%
	\includegraphics[width=0.46\linewidth]{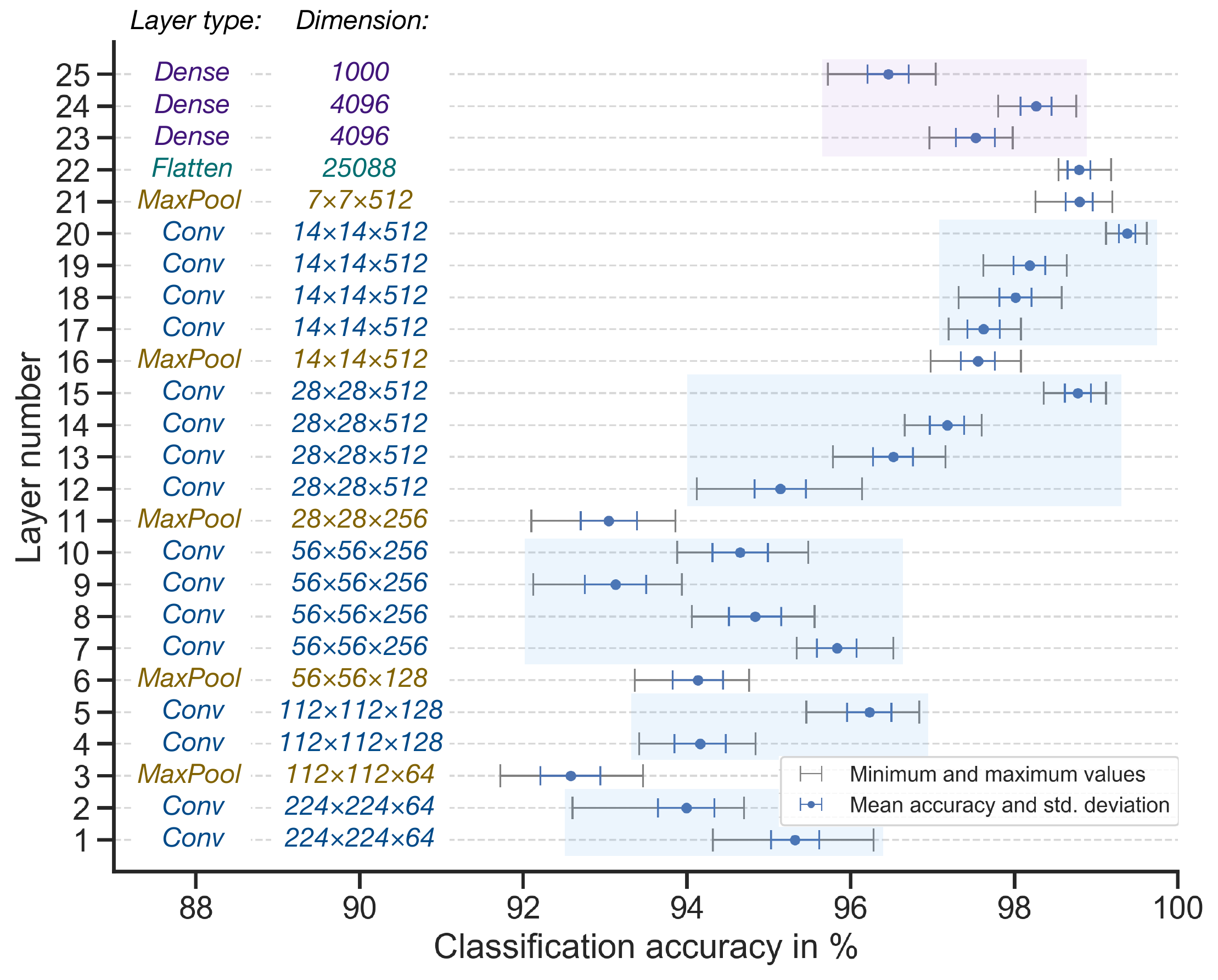}}
	
	\caption{Results of the layer-wise bit flip fault injection experiments using InjectTF2
	  and a fixed fault injection probability of $100\,\%$. The networks used are
      pre-trained VGG16 and VGG19 CNN's with ImageNet weights.}
	\label{fig:results:vgg:vgg1619}
\end{figure*}

As in the first experiment, the CMA has been computed for the tested probabilities and all layers in the network.
The sample size of this experiment can be considered as sufficient.

The results of the second category of experiments mentioned in Section~\ref{sec:experiments:layerWise} are shown in
Figure~\ref{fig:results:emil:simpleCnn:layerwise100}.
The mean value and standard deviation of the resulting classification accuracy for each layer are shown in blue alongside the minimum and maximum values that are shown in grey.
The layer numbers are shown on the Y-axis.
As before, the results have been color-coded based on the layer types, which are also listed inside Figure~\ref{fig:results:emil:simpleCnn:layerwise100} together with the corresponding output dimensions of the respective layer.

Note that the resulting accuracies are better than those in Figure~\ref{fig:results:emil:simpleCnn:layerwiseVarying}.
Because in this case, the performance is determined using the networks \emph{golden run} predictions rather than the ground truth of the dataset.
Here, the last layer of the network can be considered the most critical, since an error in this layer can directly influence the output of the network.
This is also reflected in the resulting classification accuracy.
Furthermore, the stacked convolutional layers in layers $1$\,--\,$2$ and
$5$\,--\,$6$ show a recurring pattern, which raises the question of whether the position of individual layers in relation to each other and their parameters determine the reliability of the neural network.

As expected, faults in Dropout and Flatten layers have the same drop in
classification accuracy as their preceding layer, since those layer types are not used during inference or do not influence the values of the previous layer.

Again, the CMA has been computed to verify the experiment, with a similar result as in the previous experiments.


Figure~\ref{fig:results:vgg:vgg1619} depicts the resulting classification accuracies for the layer-wise bit flip injection experiments on the VGG networks.
For each layer, the average value and corresponding standard deviation are shown in blue alongside the minimum and maximum values that are shown in grey.
The layer numbers are shown on the Y-axes.
The results have been color-coded based on the layer types, which are also listed inside Figure~\ref{fig:results:vgg:vgg1619} together with the corresponding output dimensions of the respective layer.

Overall, the performance of both networks with fault injections is similar.
Regardless of the network depth, the first layers tend to be more critical than deeper layers.
This behavior is different to our simple CNN. 
However, a general fault tolerance is observed in each case, which confirms the inherent fault tolerance of neural networks mentioned in the literature~\cite{226957, 363479, 238315}.

The first MaxPooling layers are the most critical ones in both VGG networks, as they play a key role in the extraction of feature maps.
Furthermore, it can be observed that the criticality of the convolutional layers tends to decrease with each subsequent convolutional block. 
Presumably, this is because the essential features have already been extracted in the deeper layers of the network.
The first six and last three layers in Figures~\ref{fig:results:vgg:vgg16} and \ref{fig:results:vgg:vgg19} show the same pattern in the classification accuracy drop. 
This is because the first four convolutional layers and the last three dense layers have been initialized with the same weights before training.
Only the intermediate layers were initialized randomly.

However, the plots of the VGG networks still look similar in general. Those results and the differnt behavior of our simple CNN further support the theory that the overall reliability of a neural network depends on its general architecture, the position of the layers relative to each other, and the parameters (i.e.,  the number of filters, kernel size, and output dimension) of each layer.

Also, the CMA has been computed for all layers of both the VGG16 and the VGG19 to confirm that the sample size is large enough.

\vspace{6pt}
\noindent
\textbf{Key points of the layer-wise experiments:}
\begin{itemlist}
\item The experiments help to identify the most critical layers in the network under test.
\item For our simple CNN, the last layer is the most critical one.
\item For the VGG networks, the first MaxPooling layers are the most critical ones.
\item The criticality of the convolutional layers tends to decrease with with each subsequent convolutional block.
\item Efficient selective layer-wise fault tolerance mechanisms can be introduced based on the obtained information.
\item The classification accuracy tends to drop linearly with an increasing FI probability.
\end{itemlist}


\section{Related work}
\label{sec:relatedWork}
To the best of our knowledge, only one similar FI framework for TensorFlow 1 exists.
It is TensorFI~\cite{8539213} developed by our colleagues from the University of British Columbia.
Like our InjectTF, TensorFI is capable of injecting faults into the graph of a neural network on the operation level.
The main difference between TensorFI and our approach is as follows.
TensorFI monkey patches all operations in the graph, regardless of the FI setup.
This can lead to errors when encountering TensorFlow operations that are not supported by TensorFI\@.
In contrast, InjectTF generates a new graph and changes only the operations that
are defined in the FI setup.
However, the main advantage is that our InjectTF2 supports the layer-wise fault
injection approach and TensorFlow 2.
Besides that, several other papers shall be mentioned in this Section.

Reagen et. al.\@ \cite{8465834} conduct large-scale FI experiments using their FI framework Ares.
The authors investigate the fault tolerance of various neural networks and conclude that the fault tolerance differs between different models.
This is in line with our observations from Section~\ref{sec:results:layerWise}.
In~\cite{8894493} the fault tolerance of CNNs is investigated through a series of
FI experiments. The authors focus on the injection of bit-flips
into the weights of several neural networks. They examine the impact on the
classification accuracy of the network, depending on the bit and layer position
into which bit-flips are injected.
They also propose to study the reliabiliy of a neural network based on its
architecture, which further supports our theory from Section~\ref{sec:results:layerWise}.
Schorn et. al.\@ \cite{10.1007/978-3-319-99130-6_14} propose an on-line error detection and mitigation method to detect hardware faults in NN accerators.
Using a neural network for anomaly detection, the authors are able to achieve high error detection and mitigation performace with less computational overhead than traditional redundancy-based fault tolerance approaches.
The reliability of CNNs has also been investigated by Santos et. al.\@ \cite{8536419-2}.
The authors execute CNNs for image detection on different NVIDIA GPU architectures and evaluate the impact of transient faults on the overall reliability of the network.
Based on their results on fault propagation through CNN layers, they propose a redesigned MaxPooling layer for error detection and correction.


\section{Conclusion}
\label{sec:conclusion}

In this paper we presented two new FI frameworks: \emph{InjectTF} for \mbox{TensorFlow $1$} and \emph{InjectTF2} for \mbox{TensorFlow $2$}.
In order to demonstrate the application of the frameworks, we conducted FI experiments with four different CNNs.
The CNNs have been trained using the GTSRB and ImageNet datasets.
We demonstrated how to identify the most critical operations or layers of a neural network.
These results help to introduce smart and selective fault protection mechanisms into the network.
Furthermore, the behavior of a network under the influence of faults can
be analyzed.
This allows the comparison of the fault-tolerance properties of different functionally-similar network architectures.
Based on the presented results, we proposed a theory, that the reliability of a neural network depends (i) on the overall architecture of the network, (ii) the relative position of individual layers, and (iii) the parameters (i.e., the number of filters, kernel size, and output dimension) of each layer.
The FI concepts presented in this paper have been implemented for TensorFlow.
However, both concepts can be generalized and also applied to other deep learning frameworks.
In the future, we want to investigate the proposed theory further by comparing more architectures as well as compressed, optimized networks.
Also, several improvements of the FI frameworks are planned.
First of all we will extend InjectTF2 to support networks with parallel compositions.

\bibliographystyle{Chicago}
\bibliography{bibliography}

\end{document}